\documentclass{sig-alternate}
\usepackage{url}
\usepackage{subfigure}
\usepackage[scaled=0.85]{beramono}
\usepackage{listings}
\usepackage{listings}
\newlength{\mylen}
\makeatletter
\let\@copyrightspace\relax
\makeatother
\usepackage[style=numeric]{biblatex}
\bibliography{bib.bib}

\begin{document}
\title{Leveraging Usage Data for Linked Data Movie Entity Summarization}

\numberofauthors{1} %  in this sample file, there are a *total*
\author{
% 1st. author
\alignauthor
Andreas Thalhammer, Ioan Toma, Antonio J. Roa-Valverde, Dieter Fensel\\
       \affaddr{Semantic Technology Institute}\\
       \affaddr{University of Innsbruck}\\
       \affaddr{Technikerstra{\ss}e 21a}\\
       \affaddr{6020 Innsbruck, Austria}\\
       \email{\{firstname.lastname\}@sti2.at}
}
% There's nothing stopping you putting the seventh, eighth, etc.
% author on the opening page (as the 'third row') but we ask,
% for aesthetic reasons that you place these 'additional authors'
% in the \additional authors block, viz.
% Just remember to make sure that the TOTAL number of authors
% is the number that will appear on the first page PLUS the
% number that will appear in the \additionalauthors section.
\maketitle

\begin{abstract}
Novel research in the field of Linked Data focuses on the problem of entity summarization. This field addresses the problem of ranking features according to their importance for the task of identifying a particular entity. Next to a more human friendly presentation, these summarizations can play a central role for semantic search engines and semantic recommender systems. In current approaches, it has been tried to apply entity summarization based on patterns that are inherent to the regarded data.

The proposed approach of this paper focuses on the movie domain. It utilizes usage data in order to support measuring the similarity between movie entities. Using this similarity it is possible to determine the k-nearest neighbors of an entity. This leads to the idea that features that entities share with their nearest neighbors can be considered as significant or important for these entities. Additionally, we introduce a downgrading factor (similar to TF-IDF) in order to overcome the high number of commonly occurring features. We exemplify the approach based on a movie-ratings dataset that has been linked to Freebase entities.
\end{abstract}

\category{H.1.2}{User/Machine Systems}{Human factors}[Usage Data Mining]
%A category including the fourth, optional field follows...
\category{H.3.5}{On-line Information Services}{Data sharing}[Linked Open Data]

\terms{Human Factors, Experimentation }

\keywords{linked data, entity summarization, ranking, item similarity} % NOT required for Proceedings

\section{Introduction}
\label{sec:introduction}
 Linked Data, which connects different pieces of machine-readable information (resources) via machine-readable relationships (properties) has rapidly grown in the past years, changing the way data is published and consumed on the Web. Data referring to real-world entities is being linked resulting into vast network of structured, interlinked descriptions that can be used to infer new knowledge. The rapid growth of Linked Data (LD) introduces however a set of new challenges. One in particular becomes very important when it comes to characterizing real world entities: their LD descriptions need to be processed and understood quickly and effectively. The problem known as entity summarization \cite{Cheng:2011:RRI:2063016.2063025} is concerned with identifying the most important features of lengthy LD or Linked Open Data (LOD)\footnote{\url{http://www.w3.org/wiki/SweoIG/TaskForces/CommunityProjects/LinkingOpenData}} descriptions. Solutions to this problem help applications and users of LD to quickly and effectively understand and work with the vast amount of data from LOD cloud.

In this paper we propose a novel approach that leverages usage data in order to summarize entities in the LOD space. More precisely, we perform data analysis on LD in order to identify features of entities that best characterize them. Our approach is simple and effective. We first measure similarities between entities and identify a set of nearest neighbors for each entity. For each feature of the entity we then count the number of entities having the same feature in the nearest neighbors group as well as in the set of all entities. Based on this we compute a weight for each entity, order the entities descending and select the top-n features as the summarization for each entity. To validate our approach we run a set of experiments using two datasets, namely the HetRec2011 MovieLens2k dataset \cite{Cantador:RecSys2011} and data crawled from Freebase.\footnote{\url{http://freebase.com/}} Results obtained from these datasets show that our approach is capable to identify relevant features that are shared with similar entities and thus provide meaningful summarizations.  

The remainder of this paper is organized as follows. Section \ref{sec:retrieving_important_properties} details our approach on leveraging usage data for linked data movie entity summarization. Section \ref{sec:related_work} presents the related work in the areas of entity summarization, usage mining and semantic representation of user profiles. Section \ref{sec:dataset} introduces the datasets used in our experiments while Section \ref{sec:preliminary_results} discusses the preliminary results obtained, focusing more on the neighborhood formation and neighborhood-based entity summarization results.
Finally, Section \ref{sec:conclusion} concludes the paper and Section \ref{sec:future_work} outlines future work that we plan based on the approach presented in this paper.

Please note, we use the terms item and entity interchangeable in this paper.

\section{Proposed Approach}
\label{sec:retrieving_important_properties}
\begin{figure}
\centering
\frame{
\includegraphics[]{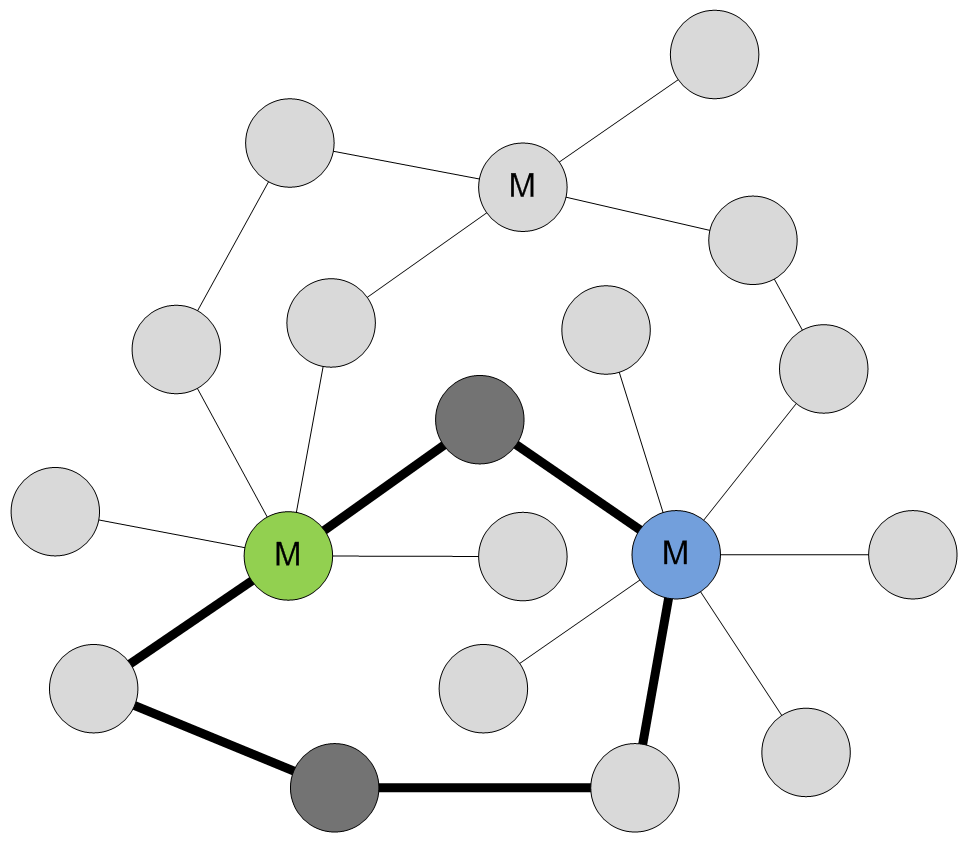}}
\caption{Visualization of shared features (strong lines and dark gray nodes) between k-nearest neighbors (green and blue nodes).}
\label{nodes1}
\end{figure}

The main idea introduced in this work is that property-value pairs - consecutively also called features - that an entity shares with its k-nearest neighbors are more relevant than features that are shared with entities that are not in the k-nearest neighbors range. Figure \ref{nodes1} visualizes this situation. Two nodes (green and blue) of the same type (M) are in each other's neighborhood. The features shared with each other (strong lines and dark gray nodes) are considered to be more important for their idendity than features they share with a node (light gray M) that is not in their respective neighborhood. The neighborhood formation of each node is based on usage data.
%We form the neighborhood (circle) for a given entity (dark gray ellipse). Light grey ellipses are of the same RDF-type as the given entity. Features that the given entity shares with at least one neighbor may be more important for the entity than others (note the different line widths).

A detailed problem statement of entity summarization is given in \cite{Cheng:2011:RRI:2063016.2063025}. The authors of this paper define the summarization of an entity $e$ as follows:
\begin{quote}
``Given $FS(e)$ and a positive integer
$k < | FS(e)|$, the problem of entity summarization is to select $Summ(e) \subset FS(e)$
such that $| Summ(e)| = k$. $Summ(e)$ is called a summary of $e$.''\footnote{In our approach, $k$ is already used for the k-nearest neighbors method. Therefore, we refer to the cardinality of the summarization as $n$.}
\end{quote}
$FS(e)$ denotes the feature set of a given entity $e$. More informally, the feature set of an entity $e$ is defined as the property-value pair set of $e$. An example for such a property-value pair for the entity $fb:en.toy\_story$\footnote{fb denotes the Freebase namespace: \url{http://rdf.freebase.com/ns/}} is:
$$\texttt{(fb:film.film.production\_companies, fb:en.pixar)}$$

In the following, $E$ denotes the set of all entities. Our approach to provide a summarization of a given entity $e \in E$ is based on usage data and includes six steps:

\begin{enumerate}
\item Generate the user-item matrix. \label{user-item}
\item Measure the similarity between $e$ and other items and identify a set $N_{k,e} \subseteq E$ of k-nearest neighbors of $e$. \label{similarity}
\item For each feature $f \in FS(e)$ collect the items $A_{e,f} \subseteq N_{k,e}$ that share the same feature. \label{neighbors}
\item For each feature $f \in FS(e)$ collect the items $B_{e,f} \subseteq E$ that share  the same feature. \label{all}
\item The weight $w$ of $f$ is the following ratio: $$w_e(f) = |A_{e,f}| \times \log \frac{|E|}{|B_{e,f}|}$$ \label{tf-idf}
\item Order the features $f \in FS(e)$ descending according to their given weight $w_e(f)$. Select the $n$ most relevant features as a summarization of $e$. \label{most-relevant}
\end{enumerate}

The concept of a user-item matrix (step \ref{user-item}) is a well-known principle in the field of recommender systems. Each column of the matrix represents a single item and each row represents a single user. The entries of the matrix are either the ratings (a numerical score) or empty if a user has not rated a particular item (which is the standard case).  The column or row vectors can be used to compare items or users amongst each other respectively. For this, several similarity measures have been introduced of which cosine similarity and Pearson correlation (comparing the vectors with regard to their angular distance) are the most common techniques \cite{Adomavicius:2005:TNG:1070611.1070751}.  

In our current implementation, we apply the \emph{log-likelihood ratio score} \cite{Dunning93accuratemethods} for computing item similarity (step \ref{similarity}). In the context of item similarity, the ratio takes into account four parameters: the number of users who rated both items, the number of users who rated the first but not the second item and vice versa, and the number of users who rated none of the two items. Note that this similarity measure does not consider the numerical values of the ratings and therefore also works with binary data like web site visits.\footnote{This is the reason why we refer to the term ``usage data'' rather than ``rating data'': we conclude usage from the process of giving a rating. We do not consider the numerical values of the ratings.} Finally, with the similarity scores it is easy to identify a set of k-nearest neighbors (kNN) for a given item.

Listing \ref{lst:sparql} states a SPARQL\footnote{SPARQL W3C Recommendation - \url{http://www.w3.org/TR/rdf-sparql-query/}} query that is used for the retrieval of common features (property-value pairs) between the item (\texttt{fb:movie.uri}) and its 20 nearest neighbors (step \ref{neighbors}). For measuring the similarity to all items in the dataset (step \ref{all}), the same query can be executed but without line 3. For each of the two result sets, the property-value pairs can be counted by occurrence. 
The filter rule (line 7) filters out property-value pairs that stem from the given entity (\texttt{fb:movie.uri}). Additionally, we also filter out the commonality of similar nearest neighbors because those features were added in the course of applying the approach and do not contribute to the summarization of the given entity.
\begin{table}[t]
\begin{lstlisting}[captionpos=t, caption=SPARQL query: retrieving property-value pairs shared with at least one of the 20-nearest neighbors.\vspace{0.15cm}, label=lst:sparql,
   basicstyle=\ttfamily,numbers=left,numberstyle=\tiny,frame=single]
select ?p ?o where {
fb:movie.uri ?p ?o.
fb:movie.uri knn:20 ?s.
?s ?p ?o.
?s rdf:type fb:film.film.
FILTER((?s != fb:movie.uri) && (?p != knn:20))
}
\end{lstlisting}
\end{table}

In the result set of the nearest neighbors, a lot of features are frequently occurring; such as the following property-value pair:
$$\texttt{(fb:film.film.country, fb:en.united\_states)}$$
If the weighting involved only counting, features like the above would be considered as highly relevant for many movies. However, as these features do not only occur often in the neighbors set but also in the overall set, they can be downgraded (step \ref{tf-idf}). As for the downgrading technique, we use the idea of the classic information retrieval method \emph{term frequency - inverse document frequency} (TF-IDF). In our case a ``term'' is stated by a single feature and the term frequency is the frequency of the feature in the nearest neighbors set. After this step, every feature that is shared with at least one of the k-nearest neighbors has an assigned weight.

Finally, in step \ref{most-relevant}, we select the $n$ most relevant property-value pairs in accordance to their weight.

%%%%%%%%%%%%%%%%%%%%%%%%%%%%%%%%%%%%%%%%%%%%%%%%%%%%%%%%%%%%%%%%%%%%%%%%%%%%%%
\section{Related Work}
\label{sec:related_work}
In the field of entity summarization, initial work has been presented in \cite{Cheng:2011:RRI:2063016.2063025}, where an approach called  RELIN is introduced. The authors apply an adapted version of the random surfer model\footnote{See also PageRank \cite{Brin:1998:ALH:297805.297827}.} - called goal directed surfer - in order to combine informativeness and relatedness for the ranking of features. In the conclusion, it is stated that a ``user-specific notion of informativeness (...) could be implemented by leveraging user profiles or feedback'' in order to mitigate the issue of presenting summarizations that help domain experts but not average users. Our approach can be considered as a first step into this direction as it focuses on leveraging usage data for providing summarizations. Our  summarizations are not adapted to each user individually but present a consensus that has been reached by similar behavior in the past.

\cite{delbru} uses combines hierarchical link analysis with weighted link analysis. For the latter, the authors suggest to combine PageRank with a TF-IDF-related weighting scheme. In this work, usage or feedback data is not considered as an additional source of information.

In the field of recommender systems, \cite{jin-zhou-mobasher} propose an approach based on Latent Dirichlet Allocation (LDA) \cite{Blei} for discovering hidden semantic relationships between items. This includes the extraction of what is considered to be the most important feature of an item (e.g. genre: adventure). The approach is exemplified on a movie and a real estate dataset.

In the  field of user modeling, there exist several approaches for leveraging (weighted) semantic knowledge about items \cite{dai-mobasher, Symeonidis, kearney}. The approach presented in \cite{dai-mobasher} proposes an aggregated presentation of user profiles by extracting and combining the domain knowledge of different items. \cite{Symeonidis} models users and items each as a feature matrix. For feature weighting in the user profile, an adapted version of TF-IDF is introduced. In the recommendation approach, the authors form neighborhoods of users based on the user-feature matrix. \cite{kearney} introduces an impact measure that indicates the influences on user behavior by item features modeled as a domain ontology. The approach is presented with examples from the movie domain.

%%%%%%%%%%%%%%%%%%%%%%%%%%%%%%%%%%%%%%%%%%%%%%%%%%%%%%%%%%%%%%%%%%%%%%%%%%%%%%%%
\section{Dataset}
\label{sec:dataset}
\begin{table}[tb]
\centering
\caption{20-nearest neighbors: Beauty and the Beast\vspace{0.15cm}}
\begin{tabular}{|r | l|}
\hline
Score & Neighbor\\
\hline\hline
0.999	 & 	\texttt{fb:en.aladdin\_1992}\\
0.999	 & 	\texttt{fb:en.the\_lion\_king}\\
0.998	 & 	\texttt{fb:en.the\_little\_mermaid\_1989}\\
0.998	 & 	\texttt{fb:en.home\_alone}\\
0.998	 & 	\texttt{fb:en.snow\_white\_and\_the\_seven\_dwarfs}\\
0.998	 & 	\texttt{fb:en.toy\_story}\\
0.998	 & 	\texttt{fb:en.mrs\_doubtfire}\\
0.998	 & 	\texttt{fb:en.the\_mask\_1994}\\
0.998	 & 	\texttt{fb:en.e\_t\_the\_extra\_terrestrial}\\
0.998	 & 	\texttt{fb:en.a\_bugs\_life\_1998}\\
0.998	 & 	\texttt{fb:en.babe}\\
0.997	 & 	\texttt{fb:en.willy\_wonka\_the\_chocolate\_factory}\\
0.997	 & 	\texttt{fb:en.honey\_i\_shrunk\_the\_kids}\\
0.997	 & 	\texttt{fb:en.men\_in\_black\_1997}\\
0.997	 & 	\texttt{fb:en.jumanji\_1995}\\
0.997	 & 	\texttt{fb:en.batman\_forever}\\
0.997	 & 	\texttt{fb:en.toy\_story}\\
0.997	 & 	\texttt{fb:en.the\_wizard\_of\_oz}\\
0.997	 & 	\texttt{fb:en.santa\_claus\_the\_movie}\\
0.997	 & 	\texttt{fb:en.who\_framed\_roger\_rabbit}\\\hline
\end{tabular}
\label{tbl:beauty1}
\end{table}
For the preparation of first tests, we combined the usage data of the HetRec2011 MovieLens2k dataset \cite{Cantador:RecSys2011} with Freebase.\footnote{\url{http://freebase.com}} The usage dataset extends the original MovieLens10M dataset\footnote{\url{http://www.grouplens.org}} by additional metadata: directors, actors, countries, and locations have been added to the original dataset. Although this dataset already contains valuable material to perform our tests without making use of LOD (i.e. Freebase), the search space for properties and objects is very restricted. In particular, 26 properties (the four mentioned above plus 22 other properties such as the genre, year, Spanish title, rotten tomatoes\footnote{\url{http://www.rottentomatoes.com/}} rating etc.) are opposed to more than 240 Freebase properties. Also, the range in Freebase is much broader as - for example - more than 380 different genres (\texttt{fb:film.film.genre}) are covered in contrast to 20 fixed genres contained in the HetRec2011 MovieLens2k dataset.

The HetRec2011 MovieLens2k dataset includes IMDb\footnote{\url{http://www.imdb.com/}} identifiers for each movie. This makes the linking to Freebase easy as querying\footnote{Freebase uses a special query language called Metaweb Query Language (MQL).} for the IMDb identifier is simple (see listing \ref{lst:mql}). Given only this query, we were able to match more than 10000 out of 10197 movies.\footnote{Unmatched items are mostly TV series that do not match the pattern \texttt{"type"="film/film/"}.}
\begin{table}[t]
\begin{lstlisting}[captionpos=t, caption=MQL query: retrieving the Freebase identifiers given an IMDb identifier.\vspace{0.15cm}, label=lst:mql,
   basicstyle=\ttfamily,numbers=left,numberstyle=\tiny,frame=single]
{
  "id"= null,
  "imdb_id"="ttIMDb_ID",
  "type"= "/film/film"
}
\end{lstlisting}
\end{table}

For performance reasons, we crawled the RDF-XML\footnote{\url{http://www.w3.org/TR/REC-rdf-syntax/}} representation from Freebase\footnote{\url{http://rdf.freebase.com/}} and stored it to a local triple store. Using the usage data, we computed the 20-nearest neighbors for each movie and stored the results also in the triple store; like in the following example:
$$\texttt{(fb:en.pulp\_fiction, knn:20, fb:en.reservoir\_dogs)}$$

Using SPARQL queries (like in listing \ref{lst:sparql}) we are able to retrieve common properties between single movies and their neighbors. The results of first tests with this setup are discussed in the following section.
\section{Preliminary Results}
\label{sec:preliminary_results}

%%%%%%%%%%%%%%%%%%%%%%%%%%%%%%%%%%%%%%%%%%%%%%%%%%%%%%%%%%%%%%%%
\begin{table*}[t]
\centering
\caption{Top-10 features: Beauty and the Beast}
\begin{tabular}{|r | l | l|}
\hline
Score & Property & Value\\
\hline\hline
39.56	&	\texttt{fb:film.film.genre}	&	\texttt{fb:en.fantasy}\\
29.40	&	\texttt{fb:film.film.rating}	&	\texttt{fb:en.g\_usa}\\
19.23	&	\texttt{fb:film.film.production\_companies}	&	\texttt{fb:en.the\_walt\_disney\_company}\\
16.89	&	\texttt{fb:film.film.music}	&	\texttt{fb:en.howard\_ashman}\\
13.31	&	\texttt{fb:film.film.music}	&	\texttt{fb:en.alan\_menken}\\
12.86	&	\texttt{fb:film.film.subjects}	&	\texttt{fb:en.fairy\_tale}\\
9.14	&	\texttt{fb:film.film.film\_casting\_director}	&	\texttt{fb:en.albert\_tavares}\\
8.04	&	\texttt{fb:film.film.written\_by}	&	\texttt{fb:en.linda\_woolverton}\\
7.75	&	\texttt{fb:film.film.produced\_by}	&	\texttt{fb:en.don\_hahn}\\
7.30	&	\texttt{fb:film.film.genre}	&	\texttt{fb:en.costume\_drama}\\\hline
\end{tabular}
\label{tbl:beauty}
\end{table*}

\begin{table*}[t]
\centering
\caption{Top-10 features: The Naked Gun - From the Files of Police Squad!}
\begin{tabular}{|r | l | l|}
\hline
Score & Property & Value\\
\hline\hline
27.77	&	\texttt{fb:film.film.written\_by}	&	\texttt{fb:en.jim\_abrahams}\\
26.00	&	\texttt{fb:film.film.written\_by}	&	\texttt{fb:en.pat\_proft}\\
22.59	&	\texttt{fb:film.film.written\_by}	&	\texttt{fb:en.jerry\_zucker}\\
22.04	&	\texttt{fb:film.film.written\_by}	&	\texttt{fb:en.david\_zucker}\\
18.92	&	\texttt{fb:film.film.music}	&	\texttt{fb:en.ira\_newborn}\\
18.44	&	\texttt{fb:media\_common.netflix\_title.netflix\_genres}	&	\texttt{fb:en.comedy}\\
16.89	&	\texttt{fb:film.film.film\_series}	&	\texttt{fb:m.0dl08h}\\
16.38	&	\texttt{fb:film.film.featured\_film\_locations}	&	\texttt{fb:en.los\_angeles}\\
16.12	&	\texttt{fb:film.film.genre}	&	\texttt{fb:m.02kdv5l}\\
15.97	&	\texttt{fb:film.film.genre}	&	\texttt{fb:en.parody}\\\hline
\end{tabular}
\label{tbl:naked}
\end{table*}

\begin{table*}[t]
\centering
\caption{Top-10 features: Bridget Jones's Diary}
\begin{tabular}{|r | l | l|}
\hline
Score & Property & Value\\
\hline\hline
29.67	&	\texttt{fb:film.film.genre}	&	\texttt{fb:en.romantic\_comedy}\\
29.39	&	\texttt{fb:film.film.written\_by}	&	\texttt{fb:en.richard\_curtis}\\
19.40	&	\texttt{fb:film.film.country}	&	\texttt{fb:en.united\_kingdom}\\
18.43	&	\texttt{fb:film.film.film\_casting\_director}	&	\texttt{fb:en.michelle\_guish}\\
16.75	&	\texttt{fb:film.film.produced\_by}	&	\texttt{fb:en.eric\_fellner}\\
16.50	&	\texttt{fb:film.film.produced\_by}	&	\texttt{fb:en.tim\_bevan}\\
13.05	&	\texttt{fb:user.robert.default\_domain.rated\_film.ew\_rating}	&	69\\
12.79	&	\texttt{fb:film.film.film\_format}	&	\texttt{fb:en.super\_35\_mm\_film}\\
12.51	&	\texttt{fb:film.film.production\_companies}	&	\texttt{fb:en.universal\_studios}\\
9.140	&	\texttt{fb:film.film.story\_by}	&	\texttt{fb:en.helen\_fielding}\\\hline
\end{tabular}
\label{tbl:bridget}
\end{table*}
\begin{table*}[t]
\centering
\caption{Top-10 features: Pulp Fiction}
\begin{tabular}{|r | l | l|}
\hline
Score & Property & Value\\
\hline\hline
21.58	&	\texttt{fb:film.film.directed\_by}	&	\texttt{fb:en.quentin\_tarantino}\\
19.75	&	\texttt{fb:film.film.genre}	&	\texttt{fb:en.crime\_fiction}\\
19.10	&	\texttt{fb:user.robert.default\_domain.rated\_film.ew\_rating} & 92\\
16.94	&	\texttt{fb:film.film.rating}	&	\texttt{fb:en.r\_usa}\\
16.38	&	\texttt{fb:film.film.featured\_film\_locations}	&	\texttt{fb:en.los\_angeles}\\
14.12	&	\texttt{fb:film.film.written\_by}	&	\texttt{fb:en.quentin\_tarantino}\\
13.72	&	\texttt{fb:film.film.film\_collections}	&	\texttt{fb:en.afis\_100\_years\_100\_movies}\\
13.48	&	\texttt{fb:film.film.edited\_by}	&	\texttt{fb:en.sally\_menke}\\
13.31	&	\texttt{fb:film.film.film\_production\_design\_by}	&	\texttt{fb:en.david\_wasco}\\
12.39	&	\texttt{fb:film.film.produced\_by}	&	\texttt{fb:en.lawrence\_bender}\\\hline
\end{tabular}
\label{tbl:pulp}
\end{table*}

%%%%%%%%%%%%%%%%%%%%%%%%%%%%%%%%%%%%%%%%%%%%%%%%%%%%%%%%%%%%%%%%
With the created dataset, we were able to identify and rank features that connect an entity to one of their nearest neighbors. We do not plan to conduct a separate evaluation at the level of neighborhood quality but we are currently in the process of performing comparisons on the level of quality of summarizations. In this analysis, we are also conducting different similarity measures as well as estimating the optimal size of the neighborhood. At the current stage of our work, statistics for the presentation of these results have not been produced.

We will discuss our findings regarding the neighborhood formation in section \ref{subsec:neigh}. Moreover, preliminary results of the entity summarization approach are presented in section \ref{subsec:summ}.

\subsection{Neighborhood formation}
\label{subsec:neigh}
One of the most important steps is the neighborhood formation dependent solely on usage data. An example for such a neighborhood is presented in table \ref{tbl:beauty1}. In general the presented neighborhood of the movie ``Beauty and the Beast'' fits the perception of most observers and also overlaps with related movies presented in IMDb.\footnote{\url{http://www.imdb.com/title/tt0101414/}, as of February 2012} The scores presented in table \ref{tbl:beauty1} are all very close to each other and every score is also close to a perfect match (1.0). In this respect, the question arises whether the k-nearest neighbor approach makes sense with such dense scores. An alternative could be to introduce a threshold rather than just selecting a fixed amount of neighbors (e.g. all movies that have a similarity higher than 0.95). As a matter of fact, the runtime of the SPARQL queries would turn into a gambling game as it can not be decided in advance whether there are 10 or 500 neighbors that cross the threshold. Another approach to address this question would be to introduce different or additional similarity measures that improve the result set while - at the same time - widens the range of the scores. Finally, the optimal neighborhood size is still due for evaluation. As such, the current size of 20 was selected to serve for the creation of first results.

A particularity about the neighborhood is that one movie (\texttt{fb:en.toy\_story}) occurs twice in the list. This is due to the HetRec2011 MovieLens2k dataset that contains several duplicates with different identifiers. We suppose that these duplicates occur due to an automatic processing that has been conducted in the course of enriching the original MovieLens10M dataset with additional data.

\subsection{Neighborhood-based entity summarization}
\label{subsec:summ}
After the neighborhood formation step we are able to extract the 10 most important features for each entity. Tables \ref{tbl:beauty} to \ref{tbl:pulp} each provide an example for a movie entity summarization.

In general, most of the presented examples have genre as one of the strongest components. In this realm, one of the real advantages of LOD can be exemplified, i.e. data richness: genres such as ``costume drama'', ``crime fiction'' or ``parody'' are missing in the HetRec2011 MovieLens2k dataset and can not be circumscribed. It is interesting to see that the property \texttt{fb:film.film.written\_by} affects all of the presented movies. In the results, the movie ``Bridget Jones's Diary'' shares with its neighbors that the scene plays in the United Kingdom while Walt Disney as the production company is surely important for the movie ``Beauty and the Beast''. It is also worth to mention that, according to our results, ``Pulp Fiction'' is under heavy influence by its director Quentin Tarantino.

The mindful reader will surely notice that not a single actor influences the presented movies. At least ``The Naked Gun - From the Files of Police Squad'' should have as an important feature the main actor Leslie Nielsen. This is due to the fact that - in Freebase - the actors are hidden behind another node that connects movies, actors, and characters. Queries that deal with such ``two-hops-each''  relationships (see listing \ref{lst:sparql2}) are hard to resolve for triple stores and yet, we were not able to produce a result set from the triple store.\footnote{We currently employ Sesame with the Native Java Store (see also \url{http://www.openrdf.org/})} However, for the near future we consider ways to circumvent this issue that does not only affect the actor feature and also help to improve the ``hop-radius'' of such queries.

Another issue that is visible in the results is the problem of data quality and the constant evolution of the data. Newly added property-value pairs like $$\texttt{(fb:user.robert.(...).ew\_rating, 92)}$$ are shared with one or two neighbors but - at this stage  - have not been assigned to a sufficient amount of entities to be downgraded with the weighting method introduced in section \ref{sec:retrieving_important_properties}.

\begin{table}[t]
\begin{lstlisting}[captionpos=t, caption=SPARQL query: retrieving property-value pairs shared with at least one of the 20-nearest neighbors.\vspace{0.15cm}, label=lst:sparql2,
   basicstyle=\ttfamily,numbers=left,numberstyle=\tiny,frame=single]
select ?p ?q ?t where {
fb:movie.uri ?p ?o.
fb:movie.uri knn:20 ?s.
?o ?q ?t.
?s ?p ?r.
?r ?q ?t.
?s rdf:type fb:film.film.
FILTER((?s != fb:movie.uri) && (?p != knn:20))
}
\end{lstlisting}
\end{table}
\section{Conclusion}
\label{sec:conclusion}
In the following we will summarize the key findings of this early stage of research.

We have presented an approach that tries to leverage usage data in order to summarize movie entities in the LOD space. This part of Semantic Web research is connected to a variety of fields, including semantic user modeling, user interfaces, and information ranking.

The goal of our research is to provide meaningful summarizations of entities. This is the task of identifying features that ``not just represent the main themes of the original data, but rather, can best identify the underlying entity'' \cite{Cheng:2011:RRI:2063016.2063025}. Our approach can be considered as a further step to this direction. Properties such as \texttt{rdf:label} or \texttt{fb:type.object.name} are currently missing as they are usually not shared with any other entity. With regard to this issue, the approach can easily be combined with another feature ranking strategy. The question whether strong weights for features that are shared with a usage-data-based neighborhood enhance the state of the art is subject to an extensive evaluation that is currently in progress of being conducted.

Additionally, we want to discuss the fact that the presented approach is restricted to a single domain and whether it can work for multiple domains or even cross-domain. Consider a electronics web shop that includes semantic meta-information about the items to be sold. Users that search a for product that fulfills their requirements (whatever those are) provide usage data that can be used to compare two products on the basis of whether they have been browsed by a same set of users (each user has watched a set of items within a given time-frame). Utilizing this information with the proposed approach can lead to a ranked list of features that a product has (e.g. 12 mega pixels in the case of digital cameras). This may help to provide meaningful product summarizations rather than listing all features that it has. However, for data hubs like DBpedia and Freebase, filtering mechanisms (like restricting to \texttt{rdf:type} film) have to be applied for not to compare apples with pears.

%Our research on this field has just started. 
%\begin{itemize}
%\item No entity-specific properties occur (e.g. label)
%\item No actors supported currently - queries with defined property chains
%\item Data quality of Freebase
%\item discussion about how features that are shared with others help to identify
%\item Domain specificity
%\end{itemize}
\section{Future work}
\label{sec:future_work}
Considering the simplicity of our current approach and the subjective quality that has already been reached, we plan to follow this track of research. In our next contributions we plan the following enhancements:
\begin{itemize}
\item An extensive evaluation of the approach will be conducted: the analysis is will include an intrinsic as well as an extrinsic evaluation with user surveys. 
\item Features that are specific to an entity (and not shared with others) will be considered in future versions of this approach. It has to be evaluated whether usage data can help with this task.
\item The problem of intermediate nodes needs to be addressed in order to provide a scalable solution. This could be done with a fixed set of important property-value pairs (like actors and characters). Another solution would be to set up triple store indexes.
\item The ideas of diversifying the results as well as a possible adaption to user profiles and context state interesting challenges.
\item With enhanced versions of the presented approach we want to move forward to the direction of user interfaces and user interaction in the context of Linked Data; also in combination with Social Media such as Twitter and Blogs.
\end{itemize}
\section*{Acknowledgements} The research leading to these results has received funding from the European Union's Seventh Framework Programme (FP7/2007-2013) under grant agreement no. 257790.

\printbibliography
\end{document}